\documentclass[conference]{IEEEtran}
\IEEEoverridecommandlockouts      

\usepackage[table, x11names]{xcolor}
\usepackage{tabularx,booktabs} 
\usepackage[export]{adjustbox}
\usepackage[colorlinks=true,linkcolor=black,citecolor=black,urlcolor=black]{hyperref}
\usepackage{times} 
\usepackage{amsmath} 
\usepackage{amssymb}  
\usepackage{graphicx}
\usepackage{algorithm}
\usepackage{mathtools}
\usepackage{algorithm}
\usepackage{algorithmicx}
\usepackage{xcolor}
\usepackage[noend]{algpseudocode}
\usepackage[font=footnotesize]{caption}
\usepackage[font=footnotesize]{subcaption}
\usepackage{enumitem}
\usepackage{textcomp}
\usepackage{epstopdf}
\usepackage{url}
\usepackage{float}

\definecolor{brewerGreen0}{HTML}{E5F5F9}
\definecolor{brewerGreen1}{HTML}{99D8C9}
\definecolor{brewerGreen2}{HTML}{2CA25F}

\definecolor{brewerCyan0}{HTML}{ECE2F0}
\definecolor{brewerCyan1}{HTML}{A6BDDB}
\definecolor{brewerCyan2}{HTML}{1C9099}

\definecolor{brewerGrey0}{HTML}{F0F0F0}
\definecolor{brewerGrey1}{HTML}{BDBDBD}
\definecolor{brewerGrey2}{HTML}{636363}

\definecolor{revisionColor}{HTML}{0238A8} 
\definecolor{lastRevisionColor}{HTML}{CC4C02} 

\usepackage{lipsum}

\usepackage{cite}

\usepackage{balance}
\usepackage{multirow} 

\usepackage{booktabs}

\usepackage{glossaries-extra}
\setabbreviationstyle[acronym]{long-short}
\glssetcategoryattribute{acronym}{nohyperfirst}{true}

\makeglossaries

\newacronym{slam}{SLAM}{Simultaneous Localization and Mapping}
\newacronym{ba}{BA}{Bundle Adjustment}
\newacronym{sfm}{SfM}{Structure from Motion}
\newacronym{pgo}{PGO}{Pose-Graph Optimization}
\newacronym{vpr}{VPR}{Visual Place Recognition}
\newacronym{sgd}{SGD}{Stochastic Gradient Descent}
\newacronym{ils}{ILS}{Iterative Least-Squares}
\newacronym{gn}{GN}{Gauss-Newton}
\newacronym{lm}{LM}{Levenberg-Marquardt}
\newacronym{pcg}{PCG}{Preconditioned Conjugate Gradient}
\newacronym{map}{MAP}{Maximum-A-Posteriori}
\newacronym{gf}{GF}{Gaussian Filters}
\newacronym{pf}{PF}{Particle Filters}
\newacronym{sdp}{SDP}{Semi-Definite Programming}
\newacronym{vo}{VO}{Visual Odometry}
\newacronym{vio}{VIO}{Visual-Inertial Odometry}
\newacronym{lo}{LO}{LiDAR Odometry}
\newacronym{vlo}{VLO}{Visual-LiDAR Odometry}
\newacronym{imu}{IMU}{Inertial Measurement Unit}
\newacronym{ros}{ROS}{Robot Operating System}
\newacronym{ate}{ATE}{Absolute Trajectory Error}
\newacronym{ape}{APE}{Absolute Pose Error}
\newacronym{rpe}{RPE}{Relative Pose Error}
\newacronym{vins}{VINS}{Visual INertial System}
\newacronym{pod}{POD}{Plain Old Data}
\newacronym{dpc}{DPC}{Dynamic Property Container}
\newacronym{boss}{BOSS}{Basic Object Serialization System}
\newacronym{mpc}{MPC}{Model Predictive Control}
\newacronym{qr}{QR}{Quadratic Regulator}

\def\slam{\gls{slam} }

\newacronym{kslam}{K-SLAM}{Keyframe-based \slam}
\def\secref#1{Sec.~\ref{#1}}
\def\figref#1{Fig.~\ref{#1}}
\def\tabref#1{Tab.~\ref{#1}}
\def\eqref#1{Eq.~(\ref{#1})}

\def\etal{\emph{et al.}}

\usepackage{amsopn}

\newcommand{\bD}{\mathbf{D}}

\newcommand{\bC}{\mathbf{C}}

\newcommand{\bH}{\mathbf{H}}

\newcommand{\bJ}{\mathbf{J}}

\newcommand{\bbR}{\mathbb{R}}

\newcommand{\bb}{\mathbf{b}}
\newcommand{\bc}{\mathbf{c}}
\newcommand{\bd}{\mathbf{d}}
\newcommand{\be}{\mathbf{e}}

\newcommand{\bx}{\mathbf{x}}

\newcommand{\bz}{\mathbf{z}}
\newcommand{\bu}{\mathbf{u}}

\newcommand{\bh}{\mathbf{h}}

\newcommand{\bp}{\mathbf{p}}

\newcommand{\bDeltax}{\mathbf{\Delta x}}

\newcommand{\defeq}{=}

\newcommand{\bOmega}{\mathbf{\Omega}}

\DeclareMathOperator*{\argmin}{argmin}

\def\g2o{$g^2o$}
\def\se3{\mathrm{SE}(3)}
\def\t2v{\mathrm{t2v}}
\def\v2t{\mathrm{v2t}}

\newcounter{todonum}

\newcommand{\rinw}{{\mathbf{X}^{W}_{R}}}

\newcommand{\bblambda}{\boldsymbol{\lambda}}

\title{\vspace{2pt}\LARGE \bf Handling Constrained Optimization in Factor Graphs for Autonomous Navigation}
\author{Barbara Bazzana \hspace{15pt}%
  Tiziano Guadagnino \hspace{15pt}%
  Giorgio Grisetti
\thanks{Barbara Bazzana, Tiziano Guadagnino, Giorgio Grisetti are with the Department of Computer,
  Control, and Management Engineering  ``Antonio Ruberti", Sapienza University of
  Rome, Rome, Italy, Email:\,\,{\tt\footnotesize{\{bazzana, guadagnino, grisetti\}@diag.uniroma1.it}}.}
}

\begin{document}
\maketitle
\thispagestyle{empty}
\pagestyle{empty}

\begin{abstract}
Factor graphs are graphical models used to represent a wide variety of problems across robotics, such as \gls{sfm}, \gls{slam}
and calibration. Typically, at their core, they have an optimization problem whose terms only depend on a small subset of variables. Factor graph solvers exploit the locality of problems to drastically reduce the computational time of the \gls{ils} methodology. Although extremely powerful, their application is usually limited to unconstrained problems. In this paper, we model constraints
over variables within factor graphs by introducing a factor graph version of the method of Lagrange Multipliers. We show
the potential of our method by presenting a full navigation stack
based on factor graphs. Differently from standard navigation stacks, we can model both optimal control for local planning and localization with factor graphs, and solve the two problems using the standard \gls{ils} methodology.
We validate our approach in real-world autonomous navigation scenarios, comparing it with the de facto standard navigation stack implemented in ROS. Comparative experiments show that for the application at hand our system outperforms the standard nonlinear programming solver Interior-Point Optimizer (IPOPT) in runtime, while achieving similar solutions. 
\end{abstract}

\section{Introduction}
\label{sec:intro}
Factor graphs are a general graphical formalism to model several
problems.  The robotics community extensively used factor graphs to
approach estimation problems, such as \gls{sfm}, \gls{slam}
and calibration\cite{ila2017ijrr, grisetti2012iros, schonberger2016cvpr, cicco2016icra}. On the one hand, they provide a graphical representation that exposes the general structure of the problem which, if exploited, allows the design of efficient algorithms. On the other hand, factor graphs allow to naturally couple problems sharing common variables by simply joining them. The solution to the joined factor graph will be then the optimum of the coupled problem. Well-known
methodologies to solve large factor graphs use variants of Iterative Least
Squares (\gls{ils})~\cite{kuemmerle2011icra, kaess2008tro, grisetti2020solver} that do not support \emph{constrained} optimization in its original formulation. 

\begin{figure}[!ht]
		\centering
		\includegraphics[width=0.15\textwidth]{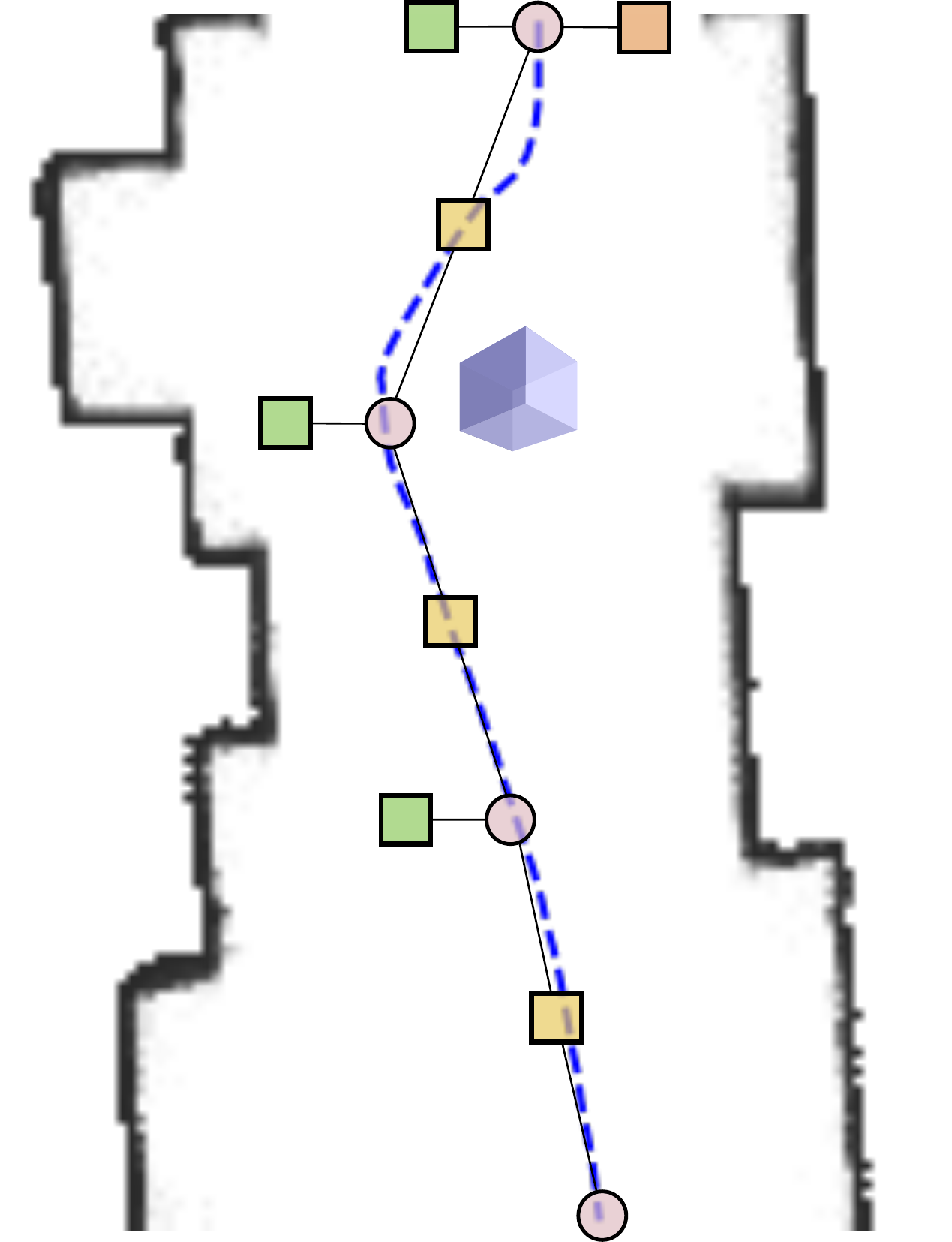}
		\hspace{15px}
		\includegraphics[width=0.2\textwidth]{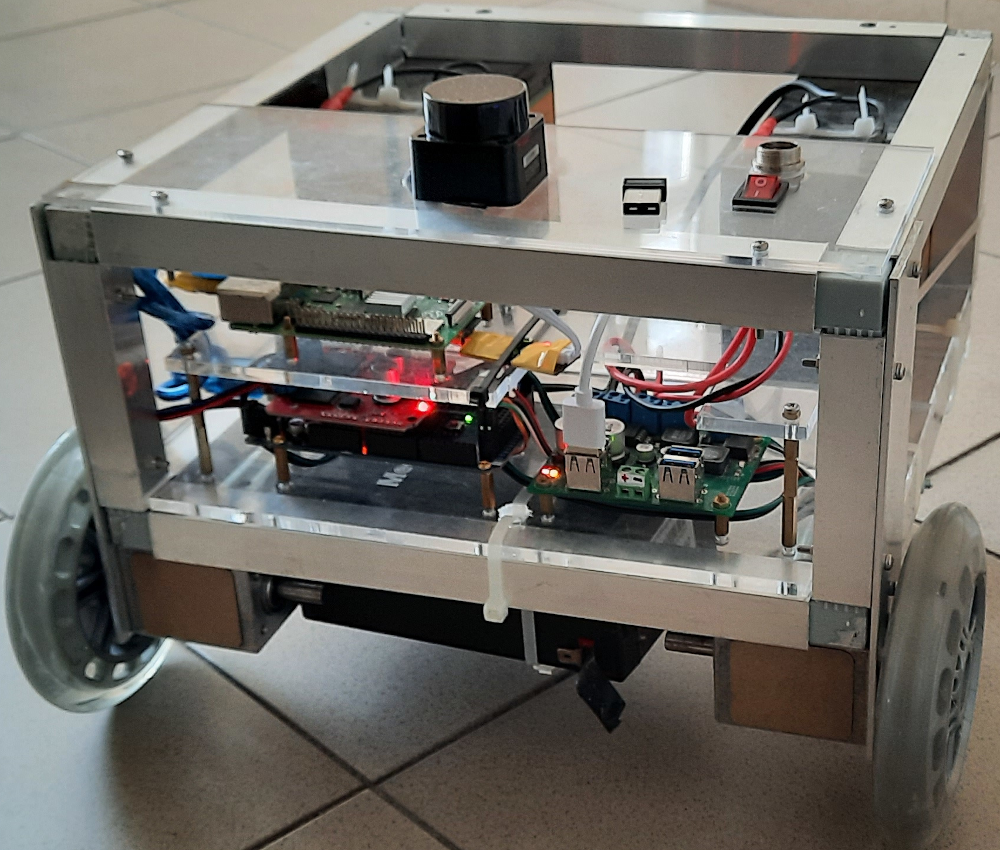}
	\caption{Left: a trajectory executed by our real robot avoiding an unforeseen obstacle, a schematic factor graph is superposed to it where some robot states are represented (pink circles). Both the localization problem (orange square) and the optimal control are depicted. States are linked by the motion model (yellow squares) and subject to obstacle avoidance and trajectory tracking (green squares). Right: our custom made unicycle.}
	\label{fig:motivation}
	\vspace{-15px}
\end{figure}

Leveraging on the results of~\cite{sodhi2020icra}, in this paper we
present an iterative version of the method of Lagrange Multipliers where constraints are modeled as a new class of factors.
Being able to perform constrained optimization allows extending the
domain of factor graphs to general optimization problems. As an application, we present here a factor graph-\gls{mpc} framework for unicycle navigation. In particular, we address here localization and motion planning within a single framework as the same general purpose factor graph solver \cite{grisetti2020solver} is used for both problems. The aim of this paper is to present a unified formulation for estimation and control problems. This unification can be very convenient as estimation and control could consistently share common information such as system dynamics, external disturbances, and uncertainty models. Thanks to the design of efficient algorithms this can be done while maintaining high computational performance.

Our
system models the localization problem as scan matching with implicit
data association where the likelihood of the current estimate is evaluated through the distance between the scan and the map.  Experimental results show that, given the odometry as initial guess, this method has similar performances to the Adaptive Monte Carlo Localization ROS package \verb+amcl+\footnote{\url{http://wiki.ros.org/amcl}}.

The optimal control problem we consider is that of generating a collision-free trajectory for a differential drive robot with actuation limits in a dynamic
environment.  To this extent, we introduce \emph{constraint} factors and \emph{obstacle avoidance} factors: the latter ones model the minimization of an artificial potential function~\cite{sfeir2011rose}\cite{farid2017iv}.


We validated our approach by running our system both in simulation and on a custom-made
robot, where all software runs on a Raspberry PI
4. Besides showing the feasibility of our approach, real-world experiments highlight its moderate computational requirements.
\figref{fig:motivation} shows a real-world trajectory and a simplistic representation of the related factor graph.
We compare our system with the well-known ROS navigation stack,
combining \verb+amcl+ for localization and the \verb+teb_local_planner+\footnote{\url{http://wiki.ros.org/teb_local_planner}}, the Timed Elastic Band~\cite{rosmann2013ecmr} ROS plugin of \verb+move_base+\footnote{\url{http://wiki.ros.org/move_base}}, for local planning. In particular, we provide a set of goals within the map of our Department and that of a factory-like environment, and consider the path length and associated duration for each segment.

Finally, experimental results confirm that when approaching constrained optimization problems on  the models we analyzed, our generic solver outperforms a state-of-the-art nonlinear programming solver such as IPOPT in runtime while producing comparable trajectories and final cost value. This gives a glimpse of the computational performances of factor graphs motivating their use also for very large problems.

\section{Related Work}\label{sec:related}

Factor graphs are a very powerful tool to model a great variety of
unconstrained optimization problems, spacing from \gls{slam} to \gls{sfm}\cite{ila2017ijrr}\cite{schonberger2016cvpr}. In
the recent literature, they have been applied to model optimal control
problems as well \cite{yang2020arxiv, ta2014icuas, xie2020corr, mukadam2019ar, dong2016rss}. To this extent, extensions of factor graph solvers have been proposed to handle constraints arising from optimal control modeling.
Formalizing optimal control problems using factor graphs allows unifying the dual problems of estimation and control under the same
representation.  Thanks to this unification, common information and variables can be shared consistently in both estimation and control processes. Further, these two aspects can be addressed jointly taking advantage of efficient factor graph solvers ~\cite{kaess2008tro, kuemmerle2011icra, grisetti2020solver}.

To the best of our knowledge, Mukadam et al.~\emph{et al.} ~\cite{mukadam2019ar} were the first to optimize a unique factor graph for both trajectory estimation and motion planning, where trajectories are modeled using continuous Gaussian Processes \cite{dong2018icra}. However, they were only relying on soft constraints. Later on, the problem of addressing constrained optimization received increasing attention, starting from Ta~\emph{et al.}~\cite{ta2014icuas}. They present a factor graph version of Sequential Quadratic Programming (SQP) to handle nonlinear equality constraints and apply their approach to the development of an \gls{mpc} framework on unmanned aerial vehicles. Our work
can be viewed as an extension of this method, where we model both equality and inequality constraints using factor graphs. In place of SQP, we rely on the Method of Lagrange Multipliers because, differently from SQP, no quadratic programming solver is required for the internal iterations when addressing inequalities.

Yang~\etal\cite{yang2020arxiv} include control inputs in a factor graph to solve a Linear Quadratic Regulator problem subject to auxiliary equality constraints. 
During the variable elimination process, when constrained variables are eliminated, a specialized solver is used for solving the constrained sub-problem separately. Differently from them, our method can handle both equality and inequality constraints and does not require a specialized solver, as it uses a standard \gls{ils} algorithm.

To the best of our knowledge, only Xie \emph{et al.}~\cite{xie2020corr} introduce both equality and inequality constraints in factor graphs. 
In particular, they present a factor graph version of a barrier-based approach to constrained optimization similar to the well-known Interior-Point Method. Our work can be viewed as complementary to Xie \emph{et al.}, in that we present an implementation of the Method of Lagrange Multipliers, as an alternative to the Interior-Point Method.


Sodhi~\emph{et. al}~\cite{sodhi2020icra} proposes an approach to refine the estimate of past trajectories: to accomplish collision-free trajectories, states are subject to inequality constraints.  To the best of our knowledge they have been the first to 
introduce the Method of Lagrange Multipliers in the context of
incremental smoothing.  We rely on the same concepts to develop a factor-graph version of this method. We propose a different application than Sodhi~\emph{et. al}: instead of state estimation, we address optimal control problems to develop a factor graph-based \gls{mpc} controller that can be used as an elegant and
compact local planner.
In addition, we address the problem
of obstacle avoidance by designing a factor that relies on
a distance cost function inspired by scan registration techniques.
The resulting factor graph is solved by our generic factor graph solver~\cite{grisetti2020solver}.

\section{Factor graph optimization}\label{sec:problem-definition}

In this paper, we model both localization and \gls{mpc} using factor graphs, which are bipartite graphs with two kinds of nodes: variables and factors. Variables represent the state of our system, while factor nodes model measurements connecting the set of variables from which they depend.


More formally, let $\bx=\bx_{0:N}$ be the set of all variables which can span over arbitrary continuous domains.  If the measurements $\bz=\bz_{0:N}$ are affected by Gaussian noise, we can represent the $k^\mathrm{th}$ factor as the tuple $\left< \bz_k, \bOmega_k, \bh_k(\cdot) \right>$  comprising the mean $\bz_k$, the information matrix $\bOmega_k$, and the prediction function $\bh(\bx_k)$, with $\bx_k$ the set of variables connected to the $k^\mathrm{th}$ factor. In this setting the negative log-likelihood of the factor graph becomes:

\begin{equation}
	\label{eq:fg_log_likelihood}
	F(\bx)=\sum_{k =0} ^{K}||\bh_k(\bx_k)-\bz_k||_{\mathbf{\Omega}_k}=\sum_{k =0} ^{K}||\be_k(\bx_k)||_{\mathbf{\Omega}_k}
\end{equation}
Solving a factor graph means finding the $\bx$ which minimizes~\eqref{eq:fg_log_likelihood}, i.e. the $\bx$ which is maximally consistent with the measurements.
In the typical case of \gls{slam}, variables can be robot states or landmark poses, while factors represent measurements correlating them. In optimal control, variables can be robot states or control inputs, whereas factors model terms of the objective function. We refer the reader to \cite{dellaert2021arcra,grisetti2020solver} for more details.  One of the aims of this paper is to address constrained optimization by extending classical \gls{ils} solvers. 

\gls{ils} minimizes \eqref{eq:fg_log_likelihood}  by using \gls{gn} algorithm that iteratively refines the current solution $\hat \bx$ by solving the quadratic approximation of \eqref{eq:fg_log_likelihood}.
The first-order Taylor expansion of the errors  $\be_k(\bx_k)$ around $\hat \bx$ given the perturbation $\bDeltax$ is computed as:
\begin{equation}
	\label{eq:fo-error}
	\be_k(\hat \bx_k + \bDeltax_k) \simeq 
	\overbrace{\be_k(\hat \bx_k)}^{\hat{\be}_k} + \bJ_k \bDeltax_k 
\end{equation}
\begin{figure}[!ht]
	\begin{center}
		\includegraphics[width=0.19\textwidth]{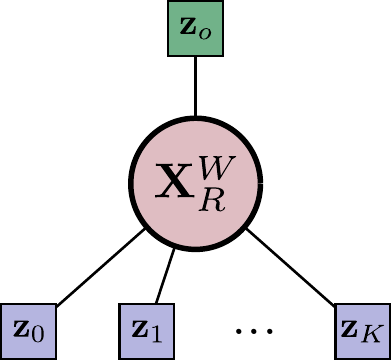}
		\caption{Factor graph modeling the localization problem. Each measurement contributes with a factor. The green square represents the odometry prior, while purple squares represent robot measurements $\bz=\bz_{0:K}$, e.g. laser endpoints.}
		\label{fig:localization-fg}
	\end{center}
	\hfill
	\vspace{-10px}
\end{figure}

\begin{figure}[!ht]
	\begin{center}
		\includegraphics[width=0.45\textwidth]{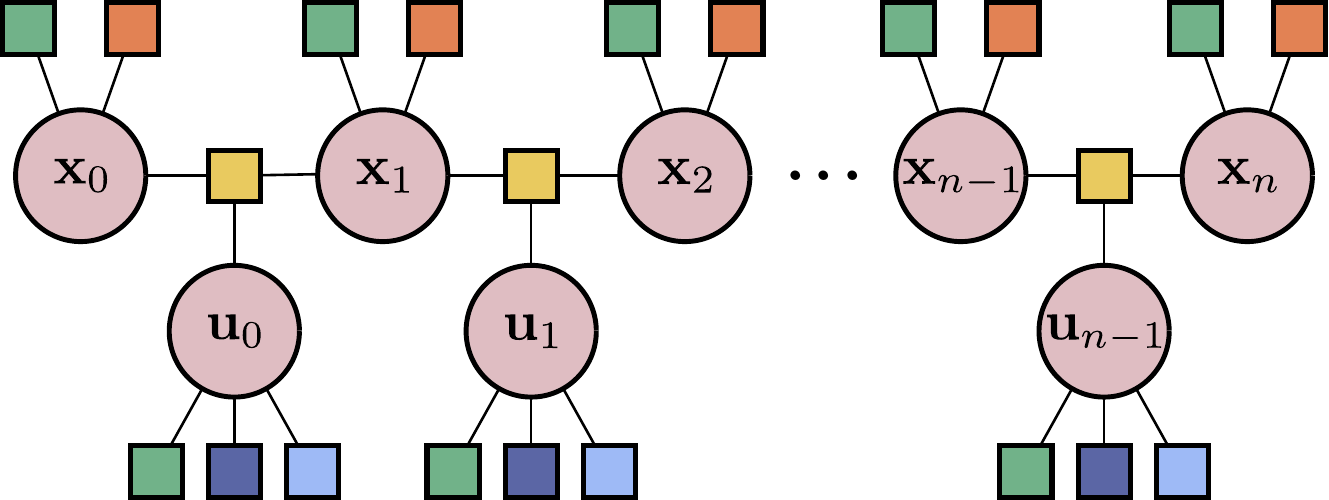}
		\caption{Factor graph modeling an optimal control problem: green squares represent prior factors imposing that the estimate is close to the desired values $\bx^{ref}$ and $\bu^{ref}$, yellow squares factors modeling the motion model, blue squares upper and lower constraint factors, orange squares obstacle avoidance factors.}
		\label{fig:optimal-control-fg}
	\end{center}
	\hfill
	\vspace{-35px}
\end{figure}

By substituting \eqref{eq:fo-error} in \eqref{eq:fg_log_likelihood}, we obtain a quadratic form that approximates the cost function around $\hat \bx$:
\begin{equation}
		\label{eq:so-approximation}
F(\hat{\bx}+\bDeltax) \simeq c+2\bb^T\bDeltax +\bDeltax^T\bH\bDeltax
\end{equation}
where  
\begin{equation}
\label{eq:linear-sys-unconstrained}
\bb=\sum_{k =0} ^{K}\bJ_k^T\bOmega_k\hat{\be}_k \qquad
\bH=\sum_{k =0} ^{K}\bJ_k^T\bOmega_k\bJ_k
\end{equation}
The minimum $\bDeltax$ of the quadratic form of \eqref{eq:so-approximation} is the solution of the linear system $\bH \bDeltax = -\bb$, while
the next estimate is updated by applying the perturbation $\hat \bx \leftarrow \hat\bx + \bDeltax$.

\section{Localization as registration on a distance matrix}\label{sec:localization}
Localization aims at estimating the most likely robot pose in the world frame $\rinw\in SE(3)$, given the robot measurements $\bz=\bz_{0:K}$ and the odometry prior $\bz_O$.
The factor graph in \figref{fig:localization-fg} illustrates this problem. In  the reminder, we will specialize this general schema in the case of laser-based localization. 
The localization module proposed leverages recent developments in photometric registration on a distance cost function $d(\cdot)$ for the laser endpoints. 

Our robot is equipped with a 2D lidar  and moves on a plane, hence $\rinw\in SE(2)$.  Let us indicate as $\bz_k\in \mathbb{R}^2$ a generic laser endpoint expressed in the robot frame. Let $d(\bp):\bbR^2 \rightarrow \bbR$ be the 2D function that returns the minimum distance between a point $\bp$ in world coordinates 
and the closest obstacle in the map. Knowing the robot pose $\rinw$ we can estimate the distance between an endpoint and the closest obstacle as:
\begin{equation}
	\be^l_k(\rinw) = d(\rinw \bz_k).
	\label{eq:loc-error}
\end{equation}
Should the robot be perfectly localized in a static environment, all distances will be close to zero, therefore \eqref{eq:loc-error} is a perfect candidate to construct a factor.
The localization graph will result in the following cost function:
\begin{align}
	G(\rinw)&= \sum_{k = 0}^{K} \|\be^l_k(\rinw\bz_k)\|_{\bOmega^l_k}\\
	\rinw^*&=\mathop{\argmin}\limits_{\rinw}G(\rinw)
\end{align}
In our implementation the distance function is computed once, and stored as a 2D matrix having the same resolution of the map.
Each cell of the matrix contains the value of $d(\bp)$ at that point. Subcell values are obtained through bilinear interpolation.
The Jacobian of \eqref{eq:loc-error} is computed by applying the chain rule where the Jacobian term originated by the distance function is computed numerically.
Besides being very intuitive, our approach combines very fast runtime with high accuracy and ease of implementation.

\section{\gls{mpc} with obstacle avoidance for unicycle}\label{sec:mpc}
As for localization, we will now describe how an optimal control problem can be represented as a factor graph.
\gls{mpc} is a well-established technique for optimally controlling systems subject to constraints. At each time step, an \gls{mpc} controller computes the best feasible trajectory which minimizes the assigned objective function, while avoiding obstacles and satisfying the constraints.  Only the first control input is applied to the system; at the next time step, a new trajectory embedding information coming from the most recent measurements is computed. More in detail, let $\bu=\bu_{0:N-1}$ be the sequence of controls, $\bx=\bx_{0:N}$ the resulting chain of states, $\bu^{ref}=\bu_{0:N-1}^{ref}$ and $\bx^{ref}=\bx_{0:N}^{ref}$ the reference values for controls and states respectively, one possible log-likelihood is the \gls{qr} cost function $\Phi_{QR}(\bu,\bx)$~\cite{bemporad2002automatica} subject to the robot dynamics and the actuation limits:
\vspace{-5px}
\begin{align}
	\label{eq:optimal-control}	
	\Phi_{QR}(\bu,\bx) = &\sum_{n=0}^{N}\|\overbrace{\bx_n-\bx_n^{ref}}^{\be^{x}_n(\bx_n)}\|_{\bOmega^x_n}
	+ \sum_{n=0}^{N-1} \|\overbrace{\bu_n-\bu_n^{ref}}^{\be^{u}_n(\bu_n)}\|_{\bOmega^u_n}
\end{align}

The link between \eqref{eq:fg_log_likelihood} and \eqref{eq:optimal-control} is evident when the control chain becomes part of the state.
As the task of \gls{mpc} is to compute both the trajectory \emph{and} the sequence of controls, in the language of factor graphs controls will become variable nodes as well.
Hence, each term in the above equation can be modeled as a prior factor of degree one, with reference value $\bx_n^{ref}$ or $\bu_n^{ref}$.  In contrast, obstacle avoidance and constrained optimization require more effort to be translated into the same formalism. \figref{fig:optimal-control-fg} illustrates the factor graph formulation of an optimal control problem.

A feasible trajectory can be found by minimizing a cost function $g(\bx)$ that decreases with the distance from obstacles, being zero at safe locations. 
According to the theory of Artificial Potential Fields~\cite{sfeir2011rose}\cite{farid2017iv}, having denoted as $k$ a positive real scalar, as $\mu$ a low-distance threshold, and as $\rho$ a high-distance threshold, we designed $g(\mathbf{x})$ as:
\begin{equation}
	\label{eq:distance-complement}
	g(\mathbf{x}) = 
	\begin{cases}
		k\bigg(\frac{1}{\mu}-\frac{1}{\rho}\bigg) & \mathrm{if} \,d(\mathbf{x}) < \mu \\
		k\bigg(\frac{1}{d(\mathbf{x})}-\frac{1}{\rho}\bigg) & \mathrm{if} \,\mu< \,d(\mathbf{x}) < \rho \\
		0 & \, \mathrm{otherwise}
	\end{cases}
\end{equation}

The above equation imposes a finite maximum value to $g(\mathbf{x})$ by clamping it to its maximum where $d(\mathbf{x}) < \mu$, and shapes a valley of zero-potential where $d(\mathbf{x}) > \rho$. Points in the region $d(\mathbf{x}) > \rho$ are safe and do not need to be pushed away by the optimization process. Therefore we model the obstacle avoidance through a factor whose error is:
\begin{equation}
	\label{eq:obs-error}
	\be^o_n(\bx_n) = g(\bx_n).
\end{equation}

It is well known that Artificial Potential Fields are affected by local minima. In our implementation, we have modified the gradient of the repulsive field generated by \eqref{eq:distance-complement} to overcome such minima. Using the idea of vortex fields ~\cite{deluca1994icra},  we add to the repulsive gradient a vector which is tangent to the equipotential contours of the repulsive field.  By applying the obstacle avoidance factor to each robot state, the vortex field acts on the whole trajectory rather than on a single state ~\cite{deluca1994icra}, to deflect it from unexpected obstacles.

Let us now concentrate on the kinematic model $\mathbf{f}(\cdot)$ mapping current state $\bx_n$ and control $\bu_n$ to next state $\bx_{n+1}$. The error can be modeled as a soft constraint:
\begin{equation}
	\label{eq:pred-error}
	\be^p_n(\bx_n) = \bx_{n+1} - \mathbf{f}(\bx_{n}, \bu_{n}).
\end{equation}
The corresponding factor has an information matrix $\bOmega^p_n$ giving the stiffness of the soft constraint. In an \gls{mpc} fashion, at each time step a graph is constructed starting from the current localization estimate which becomes the first (fixed) variable $\bx_0$.
Our robot is a unicycle, controlled in translational and rotational velocities, denoted respectively with $v_n$ and $\omega_n$.
Using Runge-Kutta integration \cite{butcher1996anm} with $T_s$ as integration interval and being $(x_n, y_n)^T$ the position and $\theta_n$ the orientation of the $n^{th}$ pose $\bx_n$, the kinematics is as follows:
\begin{align}\label{eq:kinematics}
x_{n+1} &= x_n+v_n T_s \cos\left(\theta_n + \frac{\omega_n T_s}{2}\right)\\
y_{n+1} &= y_n+v_n T_s \sin\left(\theta_n + \frac{\omega_n T_s}{2}\right)\\
\theta_{n+1} &= \theta_n + \omega_n T_s
\end{align}
Applying \eqref{eq:kinematics} in the solution scheme of \secref{sec:problem-definition} might result in controls $\bu=(v,\omega)^T_{0:N-1}$ that exceed the actuation limits.
Hence we need to enforce \emph{inequality} constraints over the variables $\bu_n$, which are not addressed by regular \gls{ils}.
In the next section, we introduce \emph{constraint factors} and \emph{constrained variables} to model both inequality and equality constraints.

Being described by the factors $\be^{x}_n(\bx_n)$, $\be^{u}_n(\bu_n)$, $\be^{o}_n(\bx_n)$, and $\be^{p}_n(\bx_n)$, the problem has a block diagonal structure. In our implementation, we impose the variable ordering $\bx_0, \bu_0, \bx_1, ...$ to guarantee a maximally sparse structure of the linear system.

\section{Extending iterative Least Squares solvers for constrained optimization}\label{sec:approach-constraint}
Lagrange Multipliers \cite{bertsekas2014constrained} is perhaps the most known method to deal with constrained optimization.
Although the method tackles nonlinear constraint functions as well, we will focus on the linear case to give a simple overview.
Consider the problem:
\begin{equation}
	\label{eq:mm}
\mathbf{x}^*=\mathop{\argmin}\limits_{\bx}\overbrace{\sum_{k=0}^{K}||\be_k(\bx_k)||_{\bOmega_k}}^{F(\bx)},\quad \mathrm{s.t.}\, \bC\bx+\bc=0
\end{equation}
where $\mathbf{C} \in \mathbb{R}^{c\times n}$.  The augmented Lagrangian of \eqref{eq:mm} is defined as:
\begin{equation}
	\mathcal{L}(\bx, \bblambda^c; \rho^c) \defeq F(\bx)+{(\bblambda^c)}^T(\bC\bx+\bc)+\frac{\rho^c}{2}||\bC\bx+\bc||^2_2.
\end{equation}
Here, $\bblambda^c\in\mathbb{R}^c$ is a vector of Lagrange Multipliers and the term $\frac{\rho^c}{2}\| \cdot \|^2_2$, with $\rho^c\in\mathbb{R},$ is a penalty term.
It can be shown \cite{boyd2011ml} that the minimum of \eqref{eq:mm} can be found by iteratively computing:
\begin{align} 
	\bx &\leftarrow \mathop {\argmin}_{\bx}\mathcal{L}(\bx, \bblambda^c; \rho^c)\label{eq:primal-update}\\
	\bblambda^c&\leftarrow\bblambda^c+\rho^c(\bC\bx+\bc)
	\label{eq:dual-update-c}.
\end{align}

This method can be generalized to deal also with inequality constraints $\bD \bx + \bd \leq 0$ with $\bD \in \mathbb{R}^{d\times n}$ by adding an additional multiplier $\bblambda^d\in\mathbb{R}^d$ and an additional penalty term weighted by $\rho^d\in\mathbb{R}$. 
The new augmented Lagrangian then becomes:
\begin{align}
	\label{eq:augmented-lagrangian}
	\mathcal{L}(\bx,\bblambda^c,\bblambda^d;\rho^c, \rho^d)=&F(\bx)\\\nonumber
	&+{(\bblambda^c)}^T(\mathbf{C}\mathbf{x}+\mathbf{c})+\frac{\rho^c}{2}||\mathbf{C}\mathbf{x}+\mathbf{c}||^2_2\\\nonumber
	&+{(\bblambda^d)}^T(\mathbf{D}\mathbf{x}+\mathbf{d})+\frac{\rho^d}{2}||\mathbf{D}\mathbf{x}+\mathbf{d}||^2_2\\\nonumber
\end{align}
The current state estimate is the $\bx$  which minimizes \eqref{eq:augmented-lagrangian}.
Further, the update-law of the inequality multiplier $\bblambda^d$ is the following:
\begin{equation}	\label{eq:dual-update-d}
\bblambda^d\leftarrow \mathrm{max}(0,\bblambda^d+\rho^d(\bD\bx+\bd))
\end{equation}
In our application, control inputs are subject to inequality constraints, namely: $|v| < v_{max}$  and $|\omega| < \omega_{max}$.

The Method of Lagrange Multipliers belongs to the class of primal-dual methods which split the optimization in two steps: a primal step (\ref{eq:primal-update}) and a dual step  (\ref{eq:dual-update-c}, \ref{eq:dual-update-d}).
Implementing the dual update requires extending the \emph{constrained variables} with the multipliers.  An iteration proceeds by updating the constrained variable $\bx$ while keeping the multipliers fixed through the primal step; once $\bx$ is refined, the multipliers $\bblambda^c$ and $\bblambda^d$ are updated with $\bx$ fixed through the dual step.  

The primal step \eqref{eq:primal-update} is a minimization that can be solved through \gls{ils} since the multipliers are fixed. Let $\hat \bx$ be the current solution, $\hat{\bc}=\bC\hat{\bx}+\bc$ and $\hat{\bd}=\bD\hat{\bx}+\bd$ be the constraint values computed at $\hat \bx$. During the linearization step of \gls{ils}, the quadratic form approximating \eqref{eq:augmented-lagrangian} around $\hat \bx$ then becomes:
\begin{equation}
	\label{eq:so-approximation-augmented}
\mathcal{L}(\hat\bx+\bDeltax,\bblambda^c,\bblambda^d;\rho^c, \rho^d) \simeq c+2\bb^T\bDeltax +\bDeltax^T\bH\bDeltax
\end{equation}

where  
\begin{align}
	\label{eq:linear-sys-constrained}
	\mathbf{b}&=\sum_{k=0}^{K}\bJ_k^T\mathbf{\Omega}_k\hat\be_k+\overbrace{\frac{\rho^c}{2}\bC^T\hat\bc\!+\!\frac{1}{2}\bC^T\bblambda^c}^{\bb^c}+\overbrace{\frac{\rho^d}{2}\bD^T\hat\bd\!+\!\frac{1}{2}\bD^T\bblambda^d}^{\bb^d}
	\\\nonumber
	\mathbf{H}&=\sum_{k=0}^{K}\mathbf{J}_k^T\mathbf{\Omega}_k\mathbf{J}_k+\overbrace{\frac{\rho^c}{2}\mathbf{C}^T\mathbf{C}}^{\bH^c}+\overbrace{\frac{\rho^d}{2}\mathbf{D}^T\mathbf{D}}^{\bH^d}.
\end{align}

By comparing \eqref{eq:linear-sys-constrained} and \eqref{eq:linear-sys-unconstrained}, it turns out that the primal update can be implemented in \gls{ils} solvers by introducing \emph{constraint factors} that contribute to the $\bH$ and the $\bb$ matrices respectively through $\bH^c$ and $\bb^c$ or $\bH^d$ and $\bb^d$, depending on the type of constraint.
Our derivation of constraint factors is made possible by the Lagrangian being a simple extension of the usual cost function. As a result, the primal update is similar to the usual update of \gls{ils}, which then triggers the update of the multipliers through the dual update. 

The addition of constraints factors $|v| < v_{max}$  and $|\omega| < \omega_{max}$ to the problem formulation presented in \secref{sec:mpc} does not change the block diagonal structure of the problem.
\section{Experiments}\label{sec:experiments}

\begin{figure}[ht]
	\begin{center}
		\includegraphics[width=0.2\textwidth, angle =90]{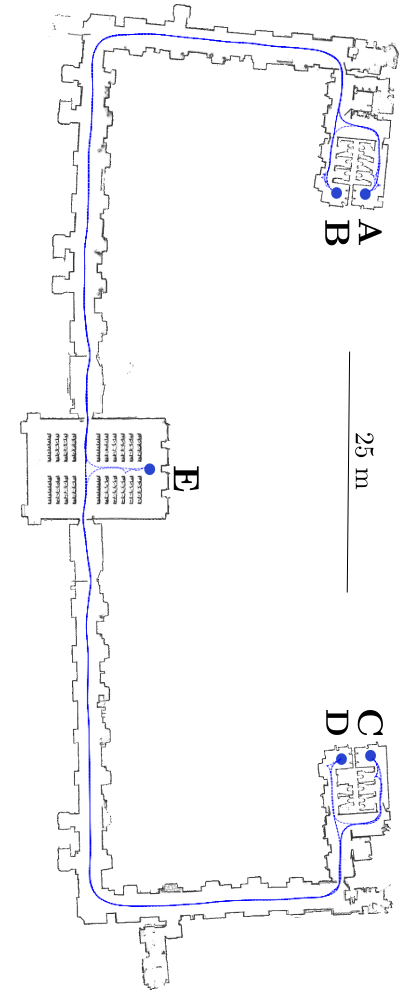}
		\caption{Map of our Department.}
		\label{fig:diag-map}
	\end{center}
	\hfill
	\vspace{-30px}
\end{figure}

We tested our system both in simulation and on a real custom made unicycle robot\footnote{https://www.marrtino.org} equipped with a 2D lidar (InnoMaker-LD06), running on a Raspberry PI4 board at 
our Department, whose map is shown in \figref{fig:diag-map}.
Comparative experiments have been conducted in simulation with a similar robot configuration, on a laptop Intel(R) Core(TM) i7-10750H CPU running at 2.60GHz with 16GB of RAM.
We evaluate separately our factor graph-based localization (\secref{sec:exp-loc-comp}) and our \gls{mpc} controller (\secref{sec:exp-mpc-comp}). For a better understanding of the behavior of \gls{mpc} optimization,
we report the internals of an \gls{mpc} iteration in \secref{sec:exp-mpc-iter}. In all simulations, we set velocity limits to $|v| <\, 1\,\mathrm{m/s}$ and $|\omega| < 1\,\mathrm{rad/s}$. 

Our navigation stack relies on 2 components: a global planner and  a local planner. Whenever a new goal is selected the global planner calculates the obstacle-free path by Dijkstra's algorithm. This path is then fed to the \gls{mpc} controller that calculates the optimal control, and considers potential dynamic obstacles detected through the lidar. The robot estimates its position by using the factor graph-based localizer of \secref{sec:localization}. 
We compared our method with the most recent version of the ROS navigation stack consisting of ROS \texttt{amcl} package for localization and ROS \texttt{move\_base} for planning and control.
\texttt{amcl} implements the Adaptive Particle Filter localization \cite{pfaff2006euros}.
As local planner for \texttt{move\_base}, we have been using the ROS package implementation of the Timed Elastic Band method\cite{rosmann2013ecmr} because the base local planner was not always able to perform the assigned navigation tasks.

\begin{figure}[ht]
	\begin{center}
		\includegraphics[width=0.5\textwidth]{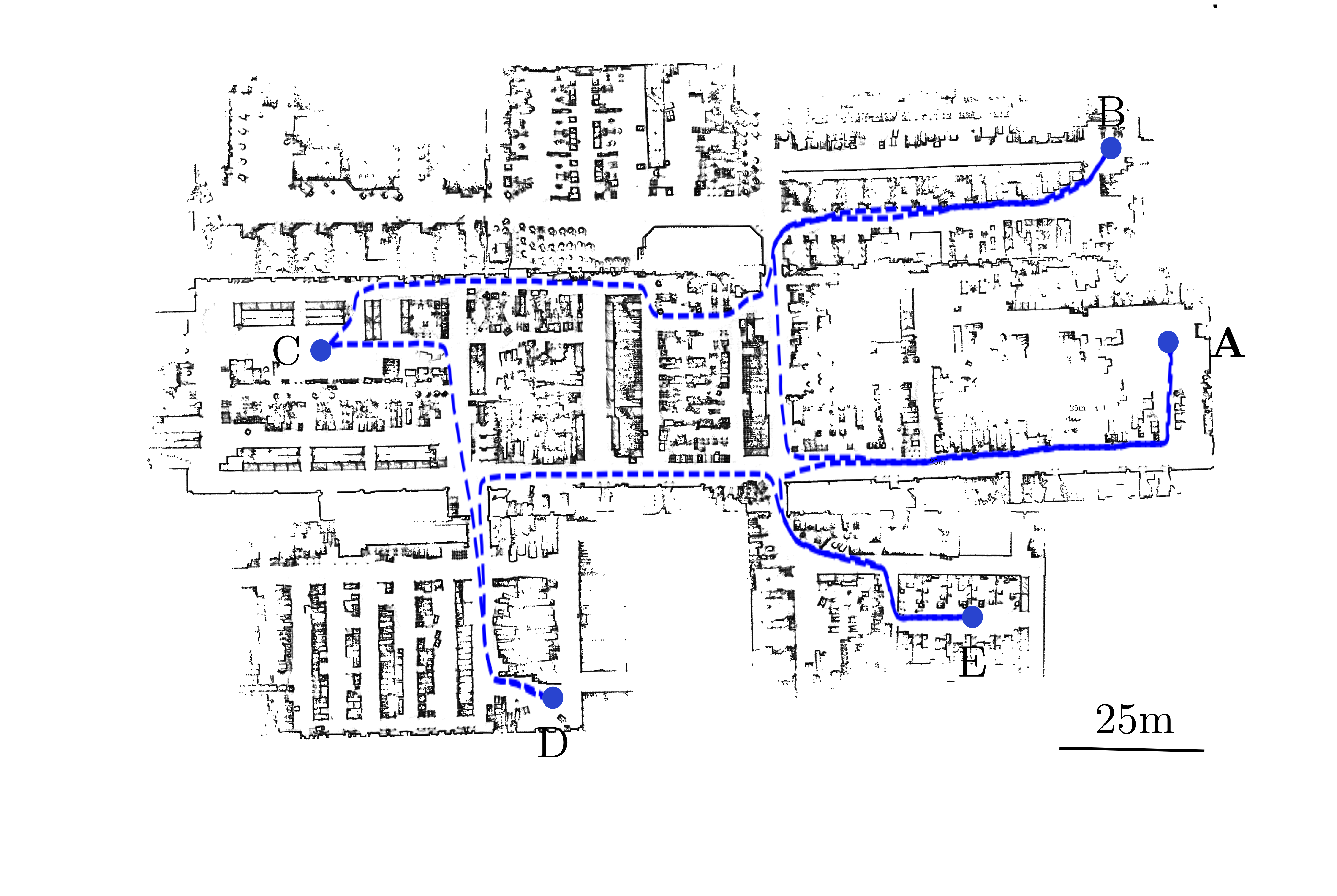}
		\caption{Map of factory-like environment.}
		\label{fig:kuka-map}
	\end{center}
	\hfill
	\vspace{-30px}
\end{figure}

In \secref{sec:exp-ipopt} we compare our method with the well-known state-of-the-art general purpose nonlinear programming solver IPOPT \cite{potra2000jcam}, which is based on Interior-Point Method. IPOPT is reliable and robust in constraint handling, with cheaper iterations than those of SQP. Our solver is on average 75 times faster while reaching a similar value of the cost function.

\begin{figure*}[ht]
	\begin{tabular}{cc}
		\includegraphics[width=0.45\textwidth]{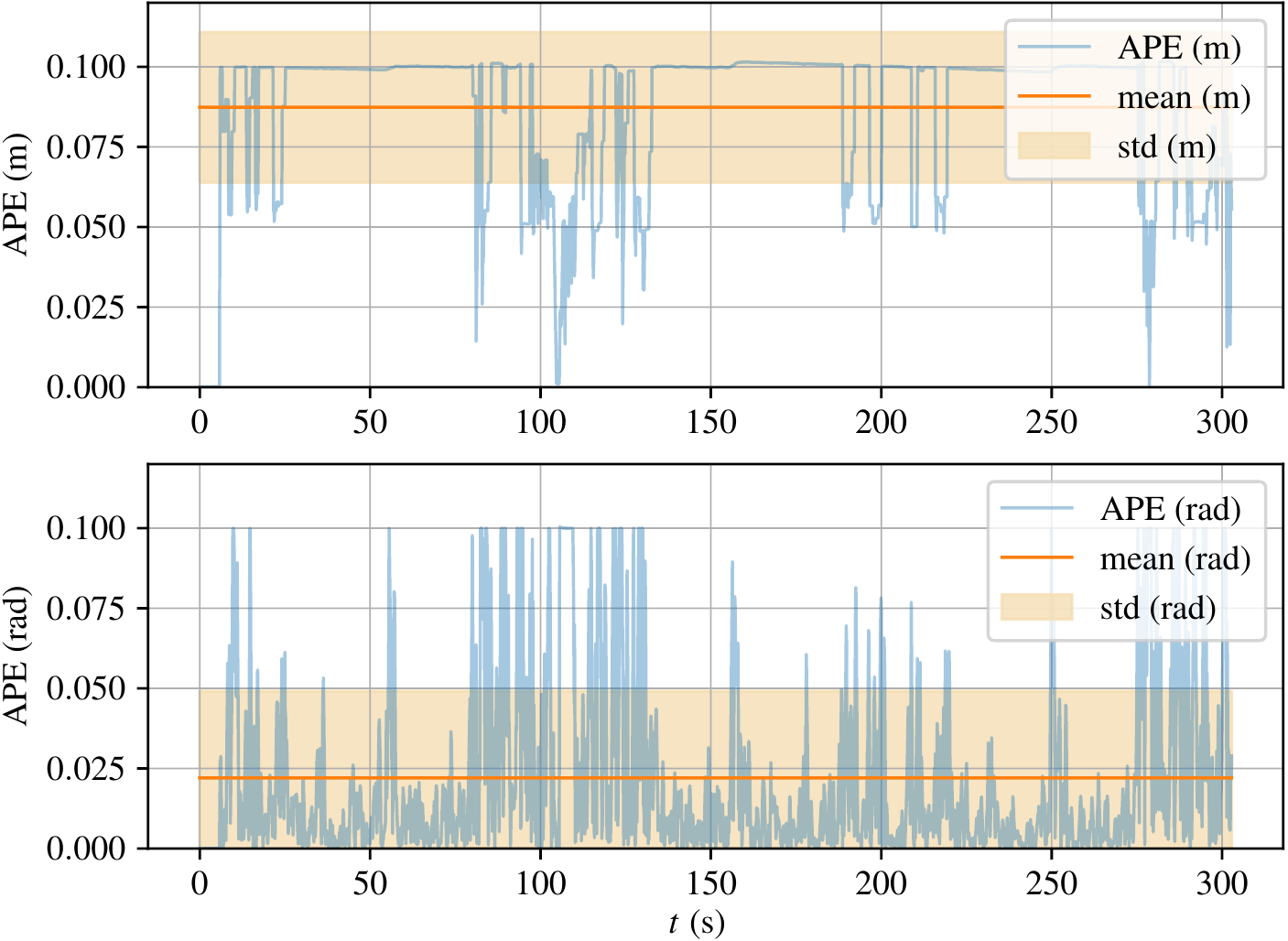}
		&
		\includegraphics[width=0.45\textwidth]{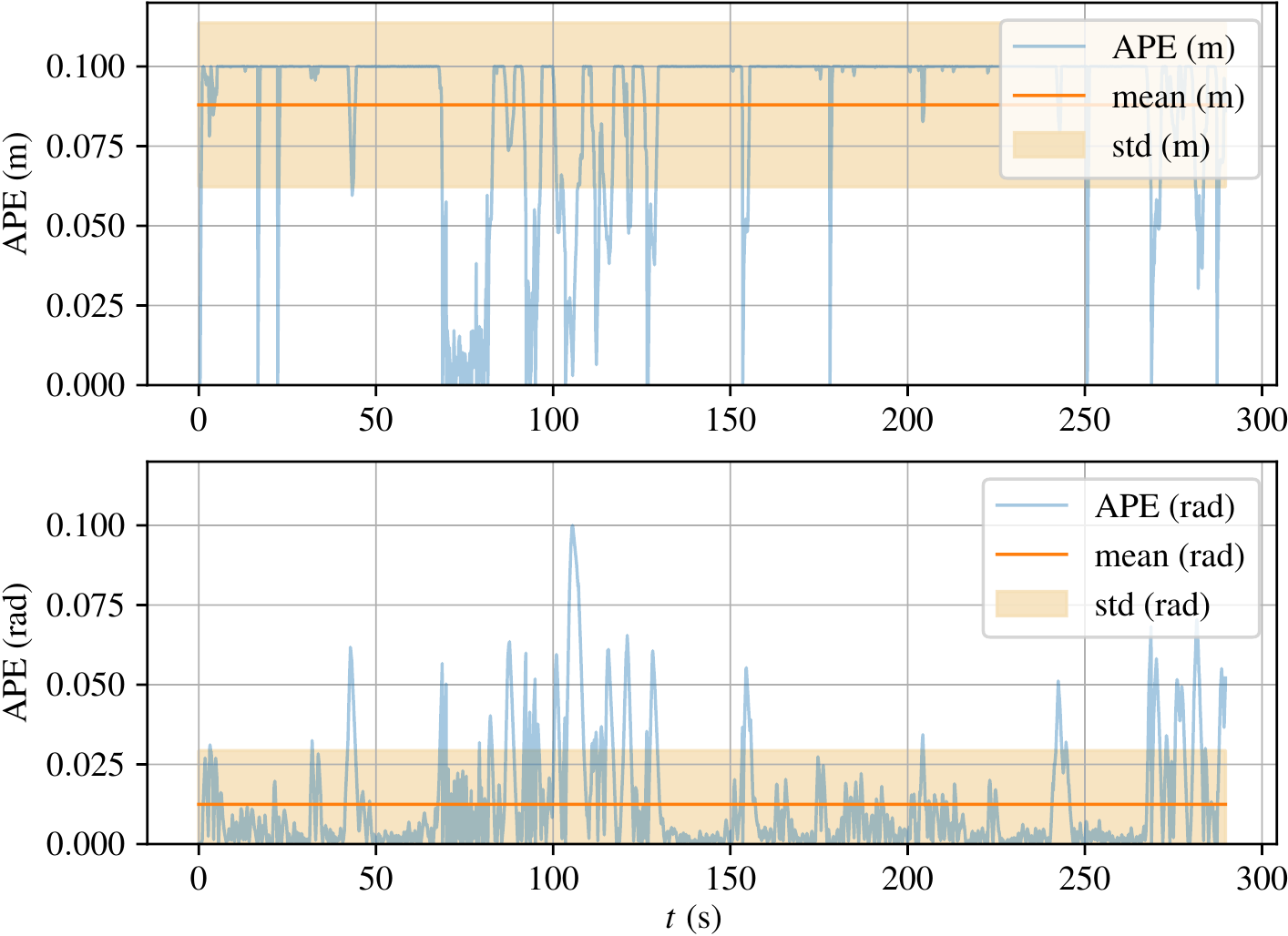}
	\end{tabular}
	\caption{APE with respect to linear and rotational part: on the left our factor graph based approach, on the right ROS \texttt{amcl}}
	\label{fig:ape}
	\vspace{-10px}
\end{figure*}

\begin{figure}[ht]
	\begin{tabularx}	{0.5\textwidth} { m{3cm} X X }
			\vspace{-12px}
			\includegraphics[width=0.37\columnwidth,height=0.32\columnwidth]{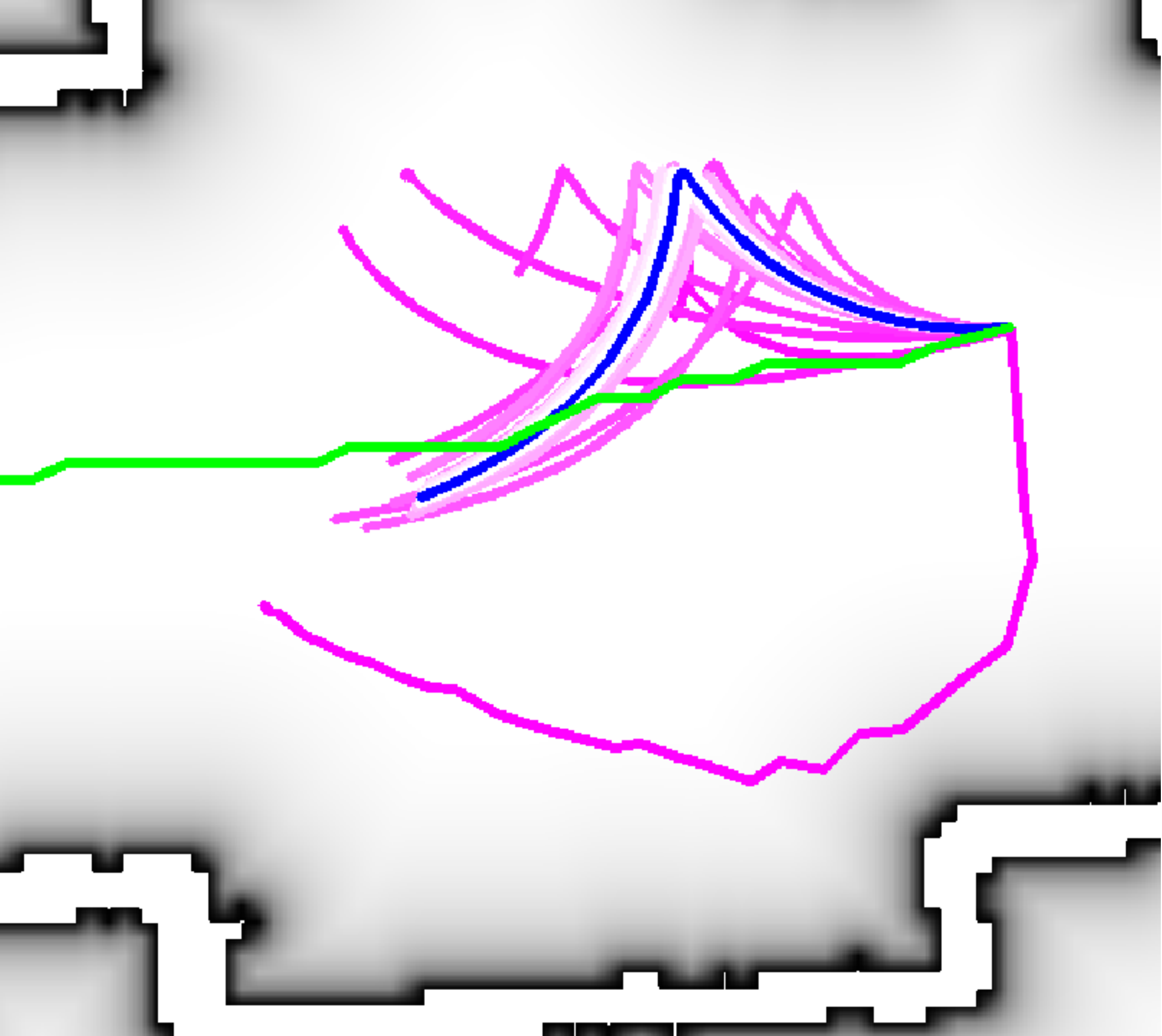}
			&
			\includegraphics[height=0.42\columnwidth,valign=c]{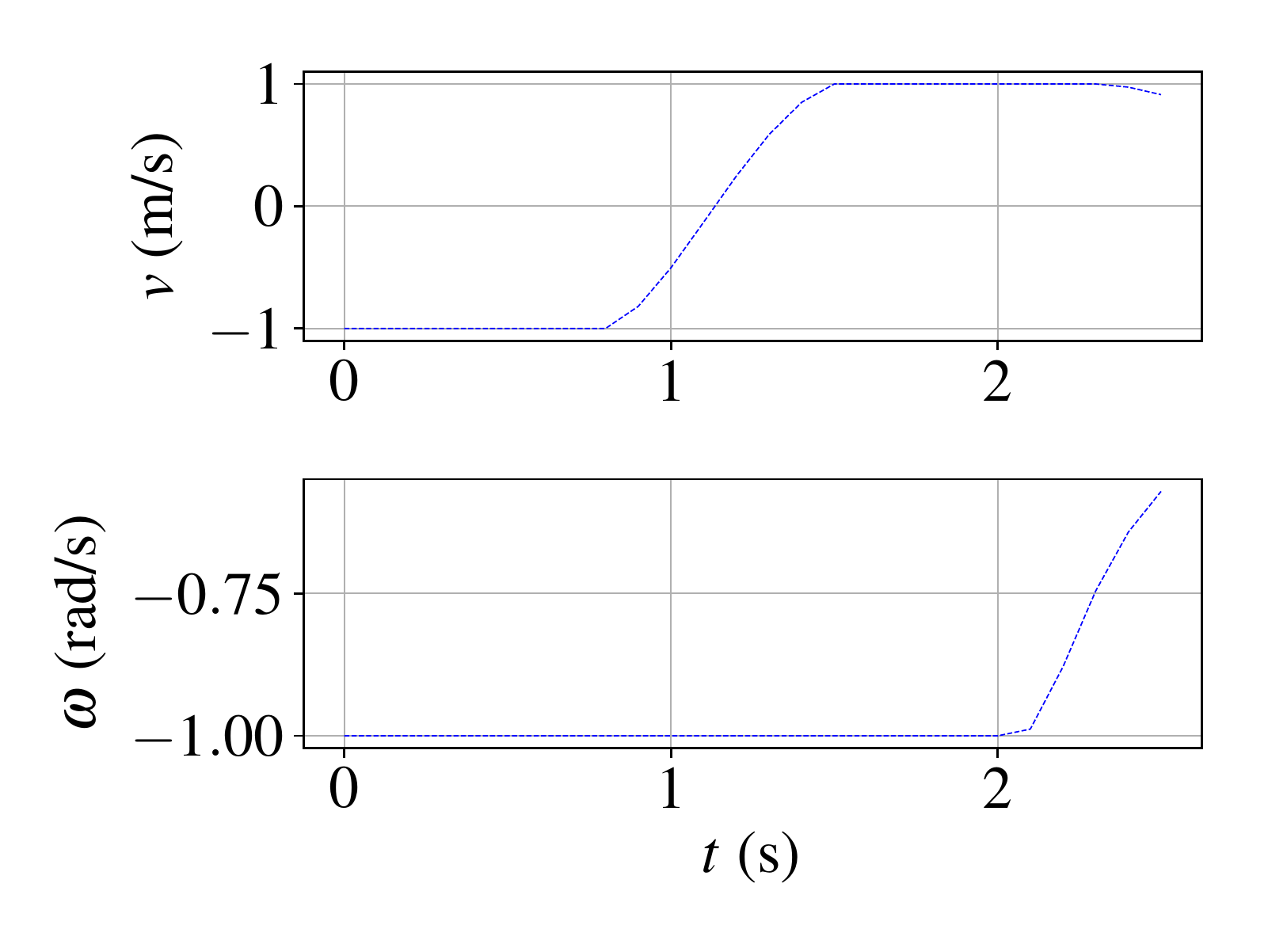}\\
			\begin{center}
				(a)
			\end{center} &     \vspace{-18px}\begin{center}
				(b)
			\end{center}\\
			\vspace{-10px}
			\includegraphics[width=0.37\columnwidth,height=0.32\columnwidth]{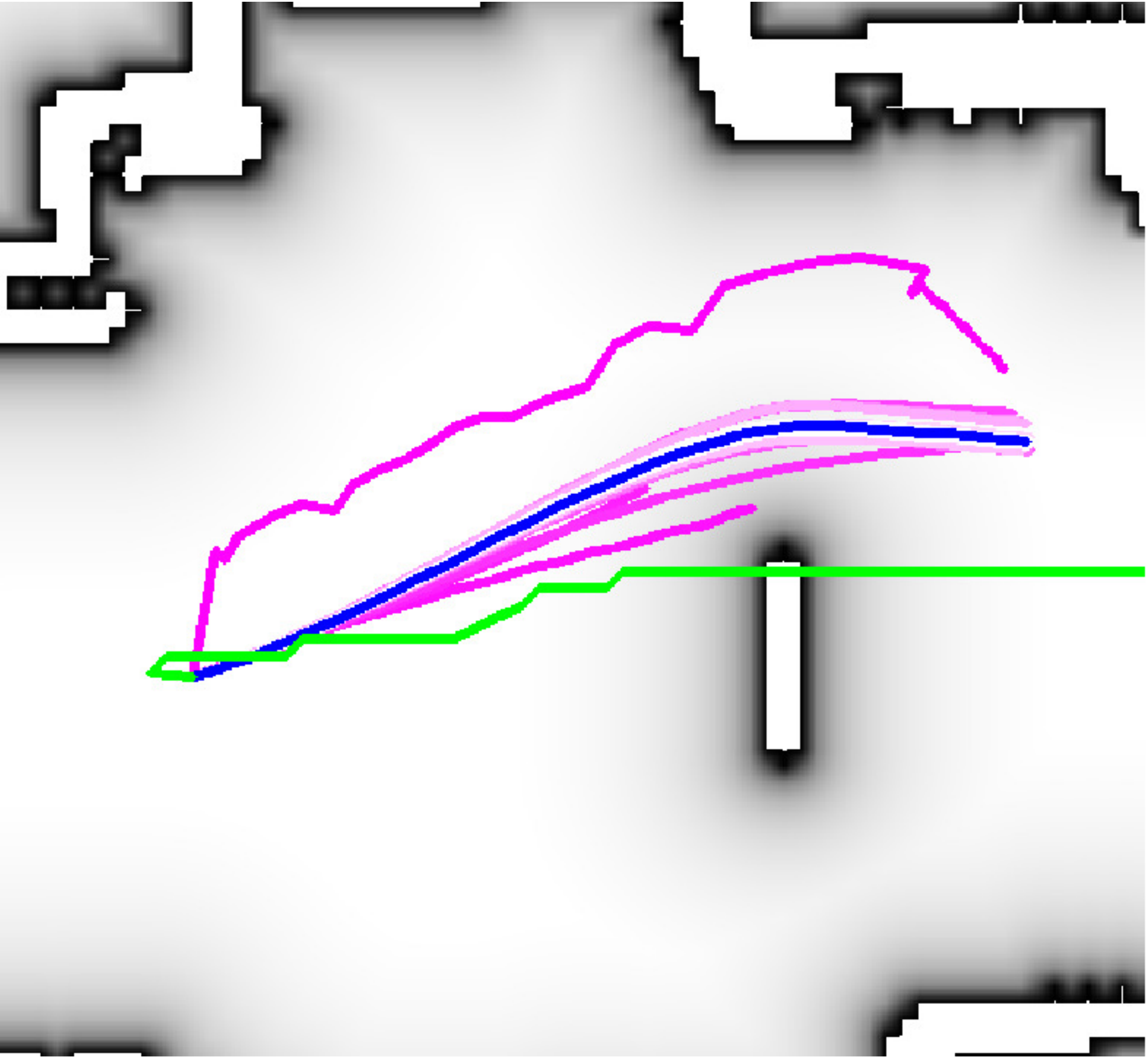} &
			\includegraphics[height=0.42\columnwidth, valign=c]{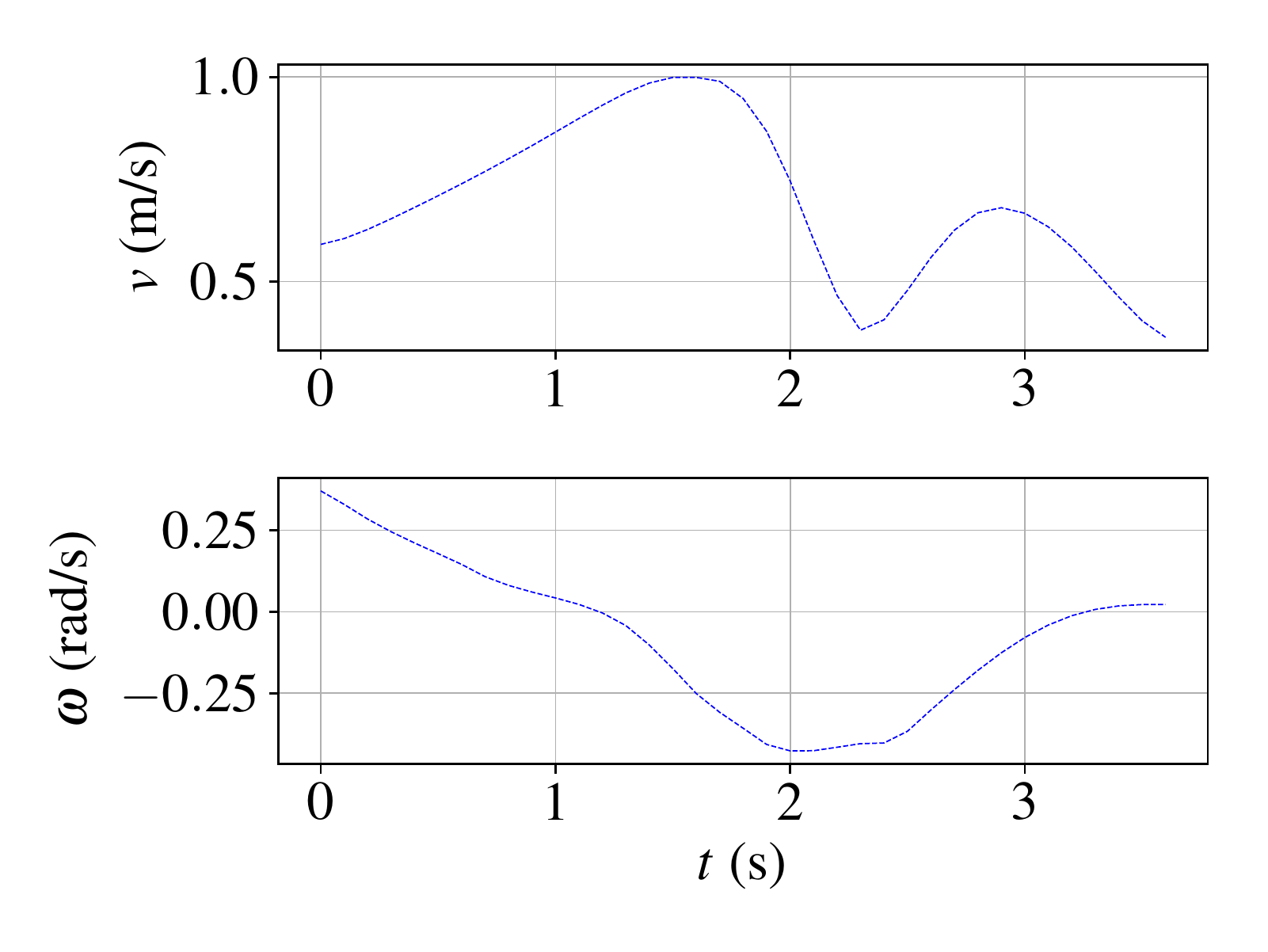}\\
			\begin{center}
				(c)
			\end{center} &     \vspace{-18px}\begin{center}
				(d)
			\end{center}
	\end{tabularx}
	\caption{Backward maneuver from right to left: (a) evolution of optimization from green curve to blue one passing through whitening pink; (b) corresponding control inputs; Obstacle avoidance maneuver from left to right: (c) evolution of optimization from green curve to blue one passing through whitening pink; (d) corresponding control inputs.}
	\vspace{-5px}
	\label{fig:maneuvers}
\end{figure}

\subsection{Localization}\label{sec:exp-loc-comp}
 We evaluated the performance of our optimization-based localization system and of \texttt{amcl} by measuring the \gls{ape} \cite{schubert2018iros} between the estimated pose and the ground truth provided by ROS \texttt{stage} environment\footnote{\url{http://wiki.ros.org/stage}}.
 In our implementation of the proposed localization method:
 \begin{itemize}
 	\item the cost of laser endpoints $\bz_k$ where $d(\rinw \bz_k)$ is smaller than the resolution of the map are scaled down by $0.25$;
 	\item laser endpoints $\bz_k$ where $d(\rinw \bz_k) \geq 0.3\,\mathrm{m}$ are discarded in the optimization to disregard the effects of unexpected obstacles;
 	\item laser endpoints $\bz_k$ further than $10.0\,\mathrm{m}$ from the robot are dropped to focus the localization effort on the local neighborhood and to reject the effect of potential distortions in the map.
 \end{itemize} 
\figref{fig:ape} plots the evolution of the rotational and translational error over time of both approaches. With an average translational error of $0.087 \pm 0.023 \,\mathrm{m}$ and rotational error of $0.022 \pm 0.027 \,\mathrm{rad}$ our approach has similar performances to \texttt{amcl} which has an error of $0.090\pm0.023\,\mathrm{m}$, $0.014\pm0.019\,\mathrm{rad}$. 

\subsection{Maneuvers through \gls{mpc}}\label{sec:exp-mpc-iter}

Our approach is capable of automatically generating control inputs that allow the robot to perform complex maneuvers. In the reminder, we will focus on two situations where the trajectories generated by our \gls{mpc} highly differ from the initial guess: backward motion and obstacle avoidance. Complex maneuvers result from satisfying the requirements expressed in the objective function while fulfilling the constraints. 

Suppose  that the robot is oriented so that the target
is behind its back. In this case, many local planning algorithms
switch to a re-orientation maneuver based on performing some kind of
backward motion until the angle between the robot orientation and the
target is below a given threshold.  Being based on optimization, our
approach will automatically generate backward maneuvers for
re-orientation, simply by minimizing the error between the desired and
the actual path. \figref{fig:maneuvers}(a) illustrates the evolution of the
optimization of our \gls{mpc} while solving for a time horizon. Thanks to
the obstacle avoidance factor, the system has the knowledge of the map
through the $g(\cdot)$ function of \secref{sec:mpc}, represented as
the background. In our implementation of \eqref{eq:distance-complement} we use $k=0.075$, $\rho=0.8\,\mathrm{m}$, and $\mu = 0.05\,\mathrm{m}$.  The optimization starts from the initial guess
illustrated by the green curve in \figref{fig:maneuvers}(a) and ends at
the blue curve. Intermediate steps are shown in pink, whitening as the iterations proceed. The
temporal evolution of the control inputs at the optimum is shown in
\figref{fig:maneuvers}(b), where the cusp corresponds to the
zero-velocity point and the limits are always met.

As we have mentioned in \secref{sec:mpc}, our optimization factor
graph outputs a feasible trajectory by imposing that it minimizes the
function $g(\cdot)$ over all poses.  \figref{fig:maneuvers}(c) highlights the
effect of the obstacle avoidance factors, using the color scheme
defined in the previous paragraph. When the initial guess coming from the
static planner crosses an unforeseen obstacle, our \gls{mpc} local planner
deflects the initial guess to overcome the obstacle while reaching the
goal.
In correspondence to the inflection point of the trajectory, the angular velocity inverts its sign while the linear velocity reaches its maximum \figref{fig:maneuvers}(d).
\begin {table}[ht]
\begin{subtable}{0.45\textwidth}
	\caption {Path length: mean $\pm$ standard deviation (m). }  
	\centering
	\begin{tabular}{ |c|c|c| } 
		\hline
		Path &Ours & ROS Navigation Stack \\
		\hline 
		AC & $175.42\pm 0.68$ ($0.39\%$)& $171.39\pm2.99$ ($1.74\%$)\\ 
		CB & $171.65\pm 0.54$ ($0.31\%$)& $169.37\pm1.09$ ($0.64\%$)\\ 
		BD & $169.52\pm 0.81$ ($0.48\%$)& $165.91\pm0.26$ ($0.16\%$)\\ 
		DE & $91.08\pm 0.37$ ($0.41\%$)& $89.66\pm0.19$ ($0.21\%$)\\ 
		EA & $93.12\pm 0.33$ ($0.35\%$)& $91.66\pm0.95$ ($1.03\%$)\\ 
		AB & $18.11\pm 0.26$ ($1.43\%$)& $18.07\pm0.21$ ($1.16\%$)\\ 
		BC & $172.76\pm 0.80$ ($0.46\%$)& $169.43\pm0.21$ ($0.12\%$)\\ 
		CD & $18.71\pm 0.19$ ($1.01\%$)& $19.07\pm0.33$ ($1.73\%$)\\ 
		\hline
	\end{tabular}
\end{subtable}
\vfill
\vspace*{10px}
\begin{subtable}{0.45\textwidth}
	\caption {Path duration: mean $\pm$ standard deviation (s)}  
	\centering
	\begin{tabular}{ |c|c|c| } 
		\hline
		Path &Ours & ROS Navigation Stack \\
		\hline 
		AC & $197.663\pm 1.63$ ($0.82\%$)& $186.13\pm4.77$ ($2.56\%$)\\ 
		CB & $193.80\pm 4.35$ ($2.24\%$)& $190.78\pm13.21$ ($6.92\%$)\\ 
		BD & $189.84\pm 1.07$ ($0.56\%$)& $176.45\pm1.06$ ($0.60\%$)\\ 
		DE & $104.43\pm 1.24$ ($1.19\%$)& $97.32\pm0.87$ ($0.89\%$)\\ 
		EA & $108.74\pm 1.37$ ($1.26\%$)& $108.46\pm13.00$ ($11.99\%$)\\ 
		AB & $31.59\pm 1.18$ ($3.74\%$)& $25.73\pm0.78$ ($3.03\%$)\\ 
		BC & $194.11\pm 3.57$ ($1.84\%$)& $182.08\pm1.46$ ($0.80\%$)\\ 
		CD & $33.67\pm 1.14$ ($3.39\%$)& $27.31\pm0.47$ ($1.72\%$)\\ 
		\hline
	\end{tabular}
\end{subtable}
\caption{Statistical analysis of path length and path duration on DIAG map.}
\label{tab:statistics}
\end{table}

\begin {table}[ht]
\begin{subtable}{0.45\textwidth}
\caption {Path length: mean $\pm$ standard deviation (m)}  
\centering
\begin{tabular}{ |c|c|c| } 
	\hline
	Path &Ours & ROS Navigation Stack \\
	\hline 
	AB & $154.30\pm 1.03$ ($0.67\%$)& $148.47\pm0.27$ ($0.18\%$)\\ 
	BC & $86.52\pm 0.31$ ($0.35\%$)& $83.99\pm0.18$ ($0.21\%$)\\ 
	CD & $142.06\pm 0.76$ ($0.53\%$)& $136.97\pm0.26$ ($0.19\%$)\\ 
	DE & $130.88\pm 1.31$ ($1.00\%$)& $124.99\pm0.19$ ($0.15\%$)\\ 
	\hline
\end{tabular}
\end{subtable}
\vfill
\vspace*{10px}
\begin{subtable}{0.45\textwidth}
\caption {Path duration: mean $\pm$ standard deviation (s)}  
\centering
\begin{tabular}{ |c|c|c| } 
	\hline
	Path &Ours & ROS Navigation Stack \\
	\hline 
	AB & $182.06\pm 1.63$ ($0.89\%$)& $197.25\pm11.78$ ($5.97\%$)\\ 
	BC & $158.96\pm 1.73$ ($0.11\%$)& $172.09\pm9.16$ ($5.32\%$)\\ 
	CD & $91.22\pm 0.65$ ($0.71\%$)& $97.99\pm5.59$ ($5.70\%$)\\ 
	DE & $147.81\pm 3.00$ ($2.03\%$)& $161.27\pm10.99$ ($6.81\%$)\\ 
	\hline
\end{tabular}
\end{subtable}
\caption{Statistical analysis of path length and path duration on a factory-like map.}
\label{tab:statistics-kuka}
\vfill
\vspace*{-10px}
\end{table}
\begin{table}[ht]
	\begin{tabular}{ |c|c|c|c|c| } 
		\hline
		Solver &Initial Cost &Final Cost&Iterations &Time[s] \\
		\hline
		\begin{tabular}{c} Ours\\IPOPT\end{tabular} &7938.200&\begin{tabular}{c} 53.612\\$\mathbf{50.423}$\end{tabular}&
		\begin{tabular}{c} $\mathbf{118}$\\445\end{tabular}&\begin{tabular}{c} $\mathbf{0.011}$\\1.459\end{tabular} \\ 
		\hline
		\begin{tabular}{c} Ours\\IPOPT\end{tabular} &3367.190&
		\begin{tabular}{c} 68.682\\$\mathbf{63.744}$\end{tabular}&
		\begin{tabular}{c} $\mathbf{125}$\\809\end{tabular}&\begin{tabular}{c} $\mathbf{0.013}$\\2.497\end{tabular} \\ 
		\hline
		\begin{tabular}{c} Ours\\IPOPT\end{tabular} &6199.650&
		\begin{tabular}{c} 64.588\\$\mathbf{60.097}$\end{tabular}&
		\begin{tabular}{c} $\mathbf{158}$\\931\end{tabular}&\begin{tabular}{c} $\mathbf{0.013}$\\2.943\end{tabular} \\ 
		\hline
		\begin{tabular}{c} Ours\\IPOPT\end{tabular} &22011.100&
		\begin{tabular}{c} 447.180\\$\mathbf{443.779}$\end{tabular}&
		\begin{tabular}{c} $\mathbf{189}$\\479\end{tabular}&\begin{tabular}{c} $\mathbf{0.031}$\\1.872\end{tabular} \\ 
		\hline
		\begin{tabular}{c} Ours\\IPOPT\end{tabular} &27744.090&
		\begin{tabular}{c} 568.386\\$\mathbf{563.406}$\end{tabular}&
		\begin{tabular}{c} $\mathbf{366}$\\549\end{tabular}&\begin{tabular}{c} $\mathbf{0.055}$\\1.951\end{tabular} \\ 
		\hline
		\begin{tabular}{c} Ours\\IPOPT\end{tabular} &21341.260&
		\begin{tabular}{c} 484.154\\$\mathbf{480.467}$\end{tabular}&
		\begin{tabular}{c} $\mathbf{294}$\\439\end{tabular}&\begin{tabular}{c} $\mathbf{0.043}$\\1.682\end{tabular} \\ 
		\hline
	\end{tabular}
	\caption{Comparison with IPOPT: convergence analysis.}
	\label{tab:ipopt}
	\vfill
	\vspace*{-10px}
\end{table}
\begin{figure}[ht]
	\centering
	\includegraphics[scale=.4]{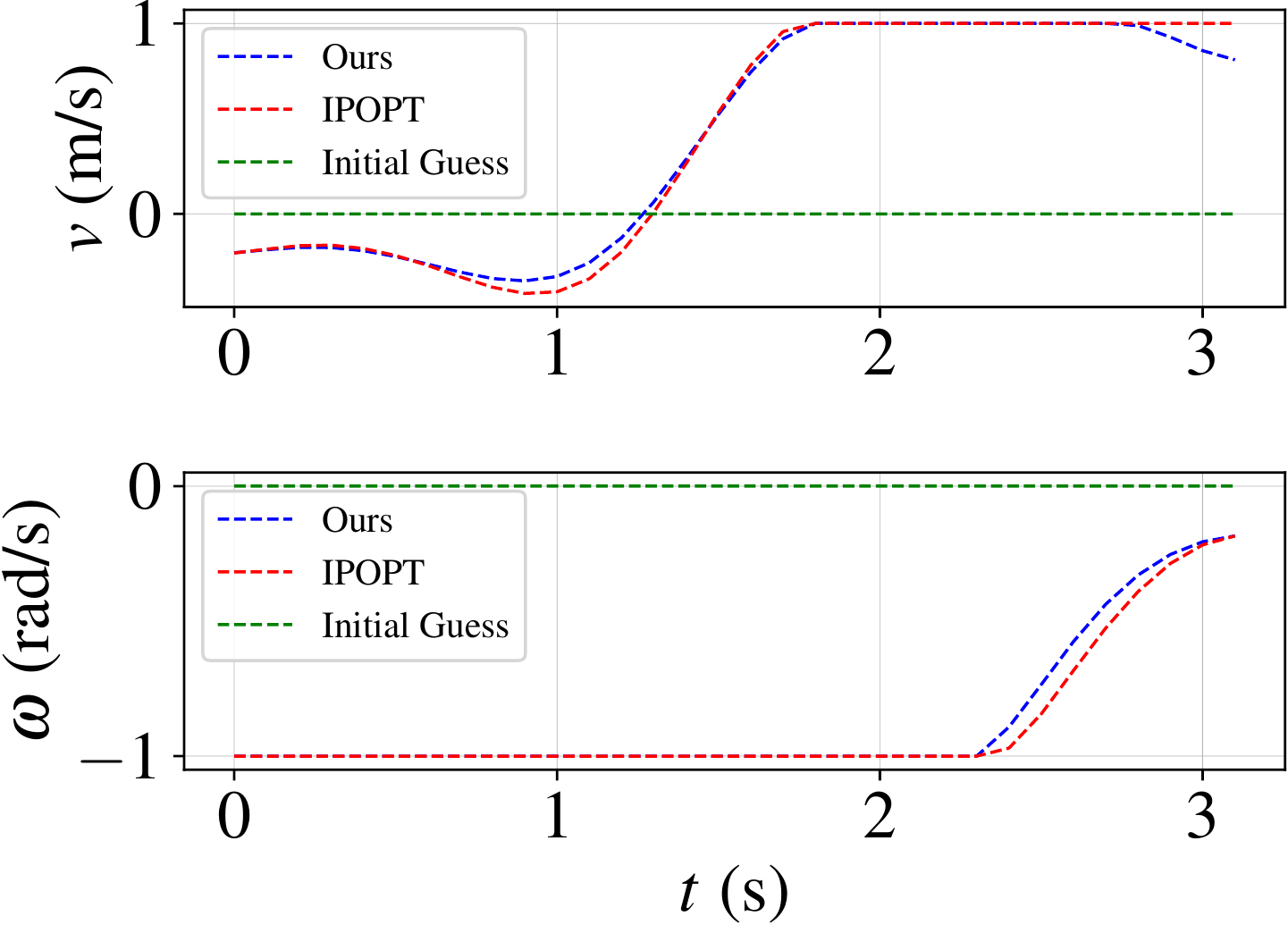}
	\caption{Comparison with IPOPT: control inputs corresponding to the backward maneuver of last row of \tabref{tab:ipopt}.}
	\label{fig:ipopt}
	\vfill
	\vspace*{-10px}
\end{figure}

\subsection{\gls{mpc} based navigation}\label{sec:exp-mpc-comp}
We report here the results obtained by running our navigation stack and the ROS one to travel across 8 goals 30 times within the map of our Department shown in \figref{fig:diag-map}. To compare the approaches we evaluated the path length and the duration to reach each of the goals. Our approach safely drove the robot during the whole 12 hours of simulation producing highly repeatable results that are summarized \tabref{tab:statistics}. For each path we store the ground truth poses published by ROS \texttt{stage} and the duration in seconds. The path length is computed by summing the distance between subsequent ground truth poses. 

Using our MPC controller, the standard deviation of path length and duration never exceed $1.43\%$ and $3.74\%$ respectively. Corresponding values obtained using the ROS navigation system are $1.74\%$ and $11.99\%$. The mean value of the percentage standard deviation on path duration is $1.88\%$ using our system and $3.56\%$ using ROS, the same value on path length is $0.60\%$ using our system and $0.85\%$ using ROS. 
On long paths, where narrow passages account only for a small portion of the whole path, ROS duration is on average $95\%$ of ours, while on paths mostly consisting of narrow passages ROS duration is $81\%$ of ours. We can infer that the difference in path duration is due to our approach begin more conservative in narrow passages.

To verify this and further validate our approach, we have tested the two navigation stacks also on the map of the factory-like environment in \figref{fig:kuka-map}. We report in \tabref{tab:statistics-kuka} the results obtained by running our navigation stack to reach 5 goals for 30 times. In this case, our approach performs better than the ROS controller in terms of path duration, still satisfying the control limits. As for the DIAG map, path duration results are more repeatable, showing a smaller standard deviation. The percentage standard deviation on path length is unchanged compared to the DIAG map results.

\subsection{Comparison with IPOPT}\label{sec:exp-ipopt}
We evaluated the performance of our solver compared to the well-known solver IPOPT. In particular, we fed an instance of our optimal control problem to the Interface For nonlinear Optimizers (IFOPT) \cite{winkler2018ifopt}, having set IPOPT as a nonlinear programming solver. Four different goals were considered starting from a given initial pose, resulting in both forward and backward trajectories. \tabref{tab:ipopt} shows the results obtained, while \figref{fig:ipopt} plots control inputs optimized with the two methods corresponding to the last row of \tabref{tab:ipopt}. As a termination criterion for our solver, we consider the norm of the perturbation vector and that of the constraint violations. Optimization is stopped when: $||\bDeltax||_{2}<\epsilon_{\bx}$, $||\bC\bx+\bc||_{\infty}<\epsilon_{\bc}$, $||max(\bD\bx+\bd, \b0)||_{\infty}<\epsilon_{\bd}$, where $\b0$ is the vector of all zeros, and $\epsilon_{\bx}, \epsilon_{\bc}, \epsilon_{\bd} = 1\mathrm{e}{-4}$. 

Our solver is around 2 orders of magnitude faster than IPOPT at converging. This makes our solver preferable for online onboard applications. Two aspects make this possible. The first is the Method of Lagrange Multipliers, whose iterations are more compact than those of Interior-Point. The second is that factor graph optimization is highly efficient for sparse optimization problems, such as the one we are solving. The final cost of our solution is slightly higher, however both a quantitative comparison with the initial value reported in \tabref{tab:ipopt} and a graphical comparison based on 	 \figref{fig:ipopt} show that the two solutions are very similar. In summary, our evaluation suggests that the proposed framework is considerably more efficient than standard non-linear programming solvers. At the same time, the proposed framework allows combining estimation and control problems under a unified methodology.

\section{Conclusions}\label{sec:conclusions}
We presented an approach to embedding constraints on variables as factors of
a graph, based on the Method of Lagrange Multipliers. Having introduced an Iterative Least Squares solver for constrained optimization, we have contributed to extending the variety of problems across robotics that can be modeled using factor graphs.  

In this paper, we focus on optimal control for \gls{mpc} as an application field of constrained optimization to support the validity of our approach. In particular, we develop a full navigation stack based on factor graphs: both localization and \gls{mpc} are modeled here through factors. Experiments show that our navigation stack compares favorably to the ROS navigation stack. Moreover, experiments conducted with IPOPT confirm that our solver gives the advantage of both an improvement in computational time and a unified formulation for estimation and control problems.  

While our navigation stack is a working proof of concept, our method
is general and can be easily used to model other classes of problems. As future work in the field of optimal control, different kinematic and dynamic models can be considered. Active SLAM ~\cite{chen2020trom, carrillo2015springer} is an example application where both estimation and control are coupled. The two problems share the knowledge of the map and the robot position estimate and contribute to the goal of having a robot that autonomously builds an accurate map in the shortest possible time. In the near future we envision future applications of our findings in this domain.

\bibliographystyle{plain}
\bibliography{glorified.bib}
\balance

\end{document}